\documentclass[runningheads,a4paper]{llncs}
\mainmatter  
\usepackage{amsmath}
\usepackage{amssymb}
\usepackage{graphicx}
\usepackage{color}
\usepackage{url}
\usepackage{hyperref}       

\makeatletter
\def\namedlabel#1#2{\begingroup
   \def\@currentlabel{#2}%
   \label{#1}\endgroup
}
\makeatother

\date{\vspace{-5ex}}

\begin{document}
\def\YHLIN#1 {{\bf x}_{{\sf lin}, \it {#1}}}

\def\YH#1 {{\bf x}_{\it {#1}}}
\def\YI#1 {{\bf u}_{\it {#1}}}
\def\YO#1 {{\bf o}_{\it {#1}}}

\def\target#1 {{\bf q}_{\it {#1}}}

\def \ti {\it t}
\def \tanh {\sf tanh}
\def \artanh {\sf artanh}

\def\WWHH {{\bf W}}
\def\WWIH {{\bf w}^{in}}
\def\WWHO {{\bf w}^{out}}

\def\itentitym {{\bf 1}}

\def\gfixp{{{\bf x}_{\infty}}}

\def \srel {\mapsto \hskip -7 pt \rightarrow }

\title
{Orthogonal Echo State Networks and stochastic evaluations of likelihoods}

\author{ 
N. Michael Mayer and  Ying-Hao Yu
}

\authorrunning{N. Michael Mayer \& Ying-Hao Yu}

\institute{
Adaptive Embedded Systems \& Robotics Laboratory \\
Department of Electrical Engineering and \\
Advanced Institute of Manufacturing with High-tech Innovations (AIM-HI),\\ 
National Chung Cheng University, Min-Hsiung, Chia-Yi, Taiwan\\
mikemayer@ccu.edu.tw
}

\maketitle

\vskip 1cm


\vfill

\definecolor{blu-gree}{rgb}{0.,.5,.25}
\centerline{\color{blu-gree} \scriptsize  \sf Preprint of a preliminary version that origins from a review process}
\centerline{\color{blu-gree} \scriptsize  \sf Final version has appeared in Cogn Comput (2017) 9:379-390
DOI 10.1007/s12559-017-9466-4.}
\centerline{\color{magenta} \scriptsize  \sf Please download the final version from}
\centerline{\url{https://link.springer.com/article/10.1007\%2Fs12559-017-9466-4}.}

\newpage

\section*{Abstract}

We report about probabilistic likelihood estimates that are performed on time series using an echo state network with orthogonal recurrent connectivity. The results from tests using synthetic stochastic input time series 
with temporal inference indicate that the capability of the network to infer depends on the balance between input strength and recurrent activity. This balance has an influence on the network with regard to the quality of inference from the short term input history versus inference that accounts for influences that date back a long time. Sensitivity of such networks against noise and the finite accuracy of  network states in the recurrent layer are investigated. In addition, a measure based on mutual information between the output time series and the reservoir is introduced. Finally, different types of recurrent connectivity are evaluated. Orthogonal matrices show the best results of all investigated connectivity types overall, but also in the way how the network performance scales with the size of the recurrent layer.

\section*{Structured abstract}
\paragraph{\bf Background.}
There is still a large gap between reservoir computing, probability theory and measures of information processing. The purpose of this paper is to provide an as simple as possible mean to calculate probabilistic likelihoods for events
 with regard to  a target function given the input history.
\paragraph{\bf Methods.} The results of the article have been achieved by symbolic and numerical analysis, as outlined in the paper.
\paragraph{\bf Results.}
The paper analytically argues for a simple way to calculate probabilistic likelihoods for a training output. As a first application this technique has been used to compute a lower limit estimate for the mutual information between a synthetic time series and a reservoir  with orthogonal recurrent connectivity and other types of recurrent connectivity. The results indicate that
the optimal performance depends on the way of balancing the input strength with the recurrent activity, which also has an influence on the network with regard to the quality of the 
inference from short term input history versus inference that accounts for influences that date back a long time in the input history. Finally, sensitivity of such networks against noise and the finite accuracy of 
network states in the recurrent layer are investigated. 
\paragraph{\bf Conclusions.} 
Methods in this paper describe and investigate in detail a potential link between reservoir computing and probabilistic modeling.
The results argue for the virtue of using orthogonal matrices for recurrent connectivity. Those reservoirs show a very satisfying performance that also scales very nicely with the number of the neurons in the recurrent layer, which is not necessarily the case for general recurrent connectivity.

\newpage

\section*{Notations in formulas}

\begin{tabular}[t]{lll}
$\WWIH$        &$\mathbb{R}^{n \times k}$         & input matrix\\
$\WWHH$        &$\mathbb{R}^{k \times k}$         & recurrent transfer matrix \\
$\WWHO$        &                                  & output matrix \\
$\YI { t } $   &$ \subset \mathbb{R}^{n}$   & input time series item\\
${\bf u}_{(-\infty,\infty)}$&                     & complete infinite input time series\\
${\bf u}_{(-\infty,t]}$&$ \in \mathbb{U}$         & input time series up to $t$\\
$\YH { t } $   &$\in \mathbb{X} \subset \mathbb{R}^{k}$    & hidden layer state time series item\\
${\bf q}_t $  & $\in \mathbb{Q} $                  & target time series \\   
$\YO { t } $       &   $\in \mathbb{O}$           & training signal for linear regression learning\\
$\tilde {\bf o}_t$ &                             & trained output of the network\\
$\omega_r  $   &                                  & an indexed subset of target values that define an event \\
$d_{r,t}$ &                                       & indicator that the event with the index $r$ has occurred \\
${\bf d}_{t}$ &                                   & vector composed of all $d_{r,t}$ \\
$<.>_{\ti}$   &                                   & mean, expectation value operator over $\ti$ \\
$s(t) = s_t $ & $\in \sf [A,B,C,D,E,F]$           & states of the test model system \\
${\cal S}_t$  &$= s_{[0,t]}$                      & time series of states until $t$\\
$ \hat{s}_t$  &                                   & best guess of the neural network w.r.t. the next state \\
$\tau,\Delta$ &                                   & internal variables of the test model system\\
$v[.]$&                                           & array of random binary numbers for the test model system\\
$a,b$&                                            & parameters for interpolations\\
\end{tabular}
\vskip 2cm

\begin{center}
In general normal type of the letters of the bold variables mentioned above are used to indicate one scalar entry in one vector, e.g.
$\YO{t} = [ o_{0,t}, o_{1,t}, \dots ]$.
\end{center}

\newpage

\section{Introduction}

The traditional approach considers the output of a neural network in the sense of a quantitative prediction, a kind of average expectancy of a trained output (e.g., backpropagation, k-means and support vector machines). 
More modern approaches cast the output in the form of probabilities and at best derive an explicit relation to some given criteria with regard to Shannon information.
The most prominent examples are 
Deep Belief Networks \cite{hinton2006fast}, but research about potential effects of stochastic resonance on neural networks \cite{bulsara1991stochastic,plesser2000noise,kingma2013auto} also refer to this field.
Only loosely related to neural networks are Bayesian networks. Bayesian networks can be used to calculate likelihoods. There inference relations have to be  known explicitly from the structure of the network, which either is implemented by hand and explicitly or it may be learned  
(e.g., by means of the K2 algorithm~\cite{chen2008improving}). Different from Bayesian networks, the present approach does not require knowledge about the structure of statistical dependencies.

One important arena of applications for neural networks, in particular recurrent neural networks (RNNs), are predictions of time series data. The idea is that any time series follows some general statistics
that allow for an estimate of the future development. Here, reservoir computing  (echo state networks (ESNs) and liquid state machines) has opened new opportunities during the last decade, which have been successfully applied in many fields (e.g. \cite{Scardapane2017}). 

So time series prediction is to infer the future values of a time series from the past, i.e., to understand its statistics.
Past values of a time series may provide significant information about future values of another time series~\cite{SHK09}. One of the possible methods for inference is transfer entropy~\cite{Sch00}. However, typical methods to measure transfer entropy have the disadvantage of requiring a fair amount of data. Granger causality~\cite{Gra69}, on the other hand, is based on regression and uses less data but is a linear method (non-linear extensions exist, see also~\cite{PQB05} for a comparison between different methods).

In this work, we propose an (non-linear) approach based on regression and a recent recurrent neural network learning method~\cite{2010c}, which we revisit in sect.~\ref{sec:esn}. Similar efforts have also been undertaken in
\cite{tivno2007markovian}. There the focus has been of a fractal representation of the input history, which is -- as a part of that work -- compared with features of a reservoir in the sense of reservoir computing. Learning is done by 
vector quantization, and the readout is performed by using the resulting tessellation of the reservoir. 

In the present work, we look into new aspects.
The first focus here is on special types of connectivity. We use orthogonal recurrent connectivity matrices. In comparison to general matrices, orthogonal matrices show special features that have proven to be particularly useful for reservoir computing \cite{boedecker2009studies,theobi2012,neco2015}.

One significant improvement with regard to orthogonal matrices versus general matrices during the numerical work for this paper is that when using orthogonal matrices the performance of the network scales nicely with the size of the hidden layer. For the approach of \cite{2010c}, general random matrices had been used which yields relatively poor results. 

The second focus is to estimate the mutual information between target time series and the reservoir, which is driven by the input time series.

Recent advances in this area have shown to be successful in time series prediction~\cite{jaeger1,jaegernips,HSS09}. However, instead of using predictions of the neural networks directly, we take a different route and describe an approach using the prediction error to detect probabilistic links in time series (sect.~\ref{sec:likelihoods}). 
Our approach is demonstrated using simulation data, first by calculating a means square error on test data (sect.~\ref{sec:MSE}), then 
by calculating a lower limit measure of mutual Shannon information between the time series and the reservoir (sect.~\ref{sec:info}). Finally, we discuss our results in sect.~\ref{sec:discussion}.

\section{Background}
\label{sec:esn}

\subsection{ESN structure}

Echo State Networks (ESNs) are an approach to address the problem of slow convergence in recurrent neural network learning. ESNs consist of three layers 
a) an input layer where the stimulus is presented to the network, b) a randomly connected recurrent hidden layer and c) the output layer. Connections in the output layer are trained to reproduce the training signal. 
It may be interpreted as a kind of interpolation that uses a recurrent random kernel \cite{e19010003}. The network dynamics are defined for discrete time-steps $\ti$, with the following equations:

\begin{eqnarray}
\YHLIN{\ti} &=& \WWHH \YH{\ti-1} +
\beta \WWIH \YI{\ti} \label{linear_response}\\
\YH{\ti} &=& 
\tanh  \left( \YHLIN {\ti} \right) 
\label{hidden_dyn} \\
\tilde{\bf o}_t &=& \WWHO \YH{\ti} \label{output1} 
\end{eqnarray}
where the vectors $\YI { \ti } $, $ \YH { \ti } $ and $\tilde{\bf o }_t$ are the input and the neurons of the 
hidden layer and output layer, respectively, and $\WWHH$ and $\WWHO$ are the matrices of the respective synaptic weight factors. $\beta$ is 
a scalar factor that controls the impact of the input by means of an independent identically distributed randomly valued input matrix $\WWIH$, where each entry is equally distributed in the range between -0.5 to 0.5. 
Input is presented to the network without additional bias. Additional bias might have an impact on the memory capacity of the network. However, this has not been investigated here. Note that the input to the network
(cf. eq. \ref{input_encoding_eq}) has values in the range between $0$ and $1$ and thus is of non-zero average input values (cf. eq. \ref{input_encoding_eq}).
Connections in the hidden layer are random. The matrix $\WWHO$ is a result of a supervised training process using a training signal $\YO{ \ti } $. 
In the following the complete sequence of $\YI{ t } $ shall be called $\YI{ (-\infty,\infty) } $.
 
\subsection{Recurrent connectivity}

In order to allow for a trainable dynamics for {\bf any} input sequence (that is a uniformly state contracting behavior), the network has to fulfill the so-called echo state property (ESP)~\cite{jaeger1,jaeger2,tighter2,e19010003}.
A network is uniformly state contracting only if 
\begin{equation}
1 \geq \max {\rm abs}(\lambda (\WWHH)), \label{necessary_eq}
\end{equation}
i.e., the largest absolute value of the eigenvalues $\lambda$  of $\WWHH$ is below $1$. A sufficient condition for a uniformly state contracting network is if there is a full rank matrix $D$ for which
\begin{equation}
1 \geq \max s (D \WWHH D^{-1}), \label{sufficient_eq}
\end{equation}
where $\max s(.)$ is the largest singular value of the matrix in the argument. 
The ESP represents an upper limit on strength of the recurrent neural network connectivity, i.e. on $\WWHH$. There are matrices $\WWHH$ that satisfy eq. \ref{necessary_eq} 
but not eq. \ref{sufficient_eq}.

In \cite{manjunath2013echo} an alternative definition for a uniformly state contracting network and ESP can be found, where a much higher connectivity strength than in eq. \ref{necessary_eq} and 
\ref{sufficient_eq} appears to be applicable. This coincides with many heuristic experiences with ESNs where many users of the networks use much higher connectivity strengths. 
However, there the ESP is always defined in relation to a particular set of input sequences. Thus, those networks do not satisfy the ESP anymore for some input sequences, especially if the value of $\beta$ is near $0$.
Different from that definition the networks that are used here are due to the general ESP that is applicable to any input sequence, which allows the features of the network, i.e. its singular value, to be chosen independent of the value of 
$\beta$. In any case the ESP is fulfilled.
This can be achieved by using matrices that have a maximal singular value that is lower or equal to  one \cite{e19010003}. We use here in all simulations recurrent matrices with a maximal singular value of one.

Normal matrices are matrices that commute with their transpose,
\[ \WWHH \WWHH^T - \WWHH^T \WWHH = 0. \]
Any normal matrix applies either to both inequalities of  eq. \ref{necessary_eq} and \ref{sufficient_eq} or to neither of them \cite{neco2015}. 

Orthogonal matrices, symmetric matrices and skew-symmetric matrices are subsets of normal matrices.  Orthogonal matrices are particularly interesting for the following reasons.
Numerical experiments show that orthogonal $\WWHH$ performs relatively efficiently in comparison to many other types of recurrent connectivity. Theoretical reasoning behind that 
has been explained in the case of linear networks \cite{PhysRevLett.92.148102}. 
Technically normal matrices have the virtue \cite{neco2015,e19010003} that all absolute eigenvalues are equal to one. Thus, transferring information by executing the linear part of the update rule (cf. eq. \ref{linear_response}) results 
in the least possible loss of information about the input history.
Recent other works point out the trade off between non-linearity and memory, e.g., \cite{verstraeten2010memory}, there also orthogonal matrices turn out to be particularly useful.

\subsection{Training caveats: Target sequence and training signal}
\label{sec:caveats}

The training set is composed of tuples
\[
T_t = \left({\bf q}_t , {\bf u}_{(-\infty,t]} \right).
\]
${\bf q}_t \in \mathbb{Q}$ is the target sequence, which is not necessarily identical to the training stimulus $\YO{ \ti } $. Rather in the scope of this paper we consider the more general definition
\begin{equation}
   \YO {\ti } = f({\bf q}_t),
\end{equation}
so the training signal is some deterministic function $f(.)$ of the target time series.

The aim of the supervised training paradigm is to reveal relations between ${\bf q}_t$ and any elements of ${\bf u}_{(-\infty,t]}$. 
As the most relevant premise to this paradigm, it is assumed that --as a minimal requirement-- the mutual information is non-zero between ${\bf q}_t$ and at least one of the values of ${\bf u}_t$. In the case of ESNs, one may formulate this minimal requirement as
\begin{equation}
0 < {\cal I}_{\bf q,u}, \label{info1} 
\end{equation}
where ${\cal I}_{\bf q,u}$ is the transfer entropy~\cite{Sch00} from the input time series to the output time series.

There are principal limitations on what the network can learn that are due to the fact that supervised learning is used and other limitations that result from technical features of the network design.

\begin{itemize}
\item[\sf (L1)\namedlabel{l1}{\sf L1}] An output sequence can only be learned as a function (i.e., a surjective or bijective mapping from the input to the output).
\item[\sf (L2)\namedlabel{l2}{\sf L2}] There are more narrow limits on a reservoir that is fed by an input time series to learn an output time series because the reservoir can only represent a 
final subset of the input history and thus always contains less information than the input history itself. 
\item[\sf (L3)\namedlabel{l3}{\sf L3}] 
There are more narrow limitations if one accounts for the fact that the reservoir training is done by linear regression. So the trained output is a linear function (not a general function) of the reservoir states.
\item[\sf (L4)\namedlabel{l4}{\sf L4}] 
In addition, each neuron of the reservoir is modeled with a finite accuracy (here double precision floating point variables with 64 bits). The performance of the network is also influenced by the limited accuracy. Single precision
variables (32 bit floating point) can further reduce the network's performance. 
\end{itemize}

Jaeger's ESP theorem \cite{jaeger1} states that in principle the network can indeed learn $\YO{\ti} $ 
by linear regression if the network is uniformly state contracting, that is, if 
\begin{itemize}
\item the ESP applies to the network and 
\item the ground truth relation $ \YO{\ti} (\YI{ t-1 } , \YI{ t-2 } , \YI{ t-3 } \dots )$ does not explicitly depend on $t$, i.e., the ground truth relation is shift-invariant in time.
\end{itemize}

Thus, the network performance depends on several factors such as the size of the network, the difficulty of the 
prediction of the target and the type of the recurrent connectivity.

Often and in the simplest case, training is done by setting 
\begin{equation}
 \YO{\ti} = {\bf q}_t. \label{av_tr}
\end{equation}

As mentioned above the basic assumption with regard to the relation between the output time series $\YO{ t } $ and ${\bf u}_{(-\infty,t]} $  is that at least they are not statistically independent from each other, i.e.  ${\cal I}_{\bf q,u}$ is more than zero. Here, requirements on the relation between the time series are much higher. In fact, in the best case an optimal output $\tilde{\bf o }_t$ 
of a trained network can be trained by some kind of target sequence $\target{\ti} $ that has a relation to the input history. Thus after training, the output may be virtually identical to the target sequence

\begin{equation}
\tilde{\bf o}_t \approx {\bf q}_t
\end{equation}
One then may call $\tilde{\bf o }_t$ the prediction of the network for the next value of the target time series with regard to the input history. 
However, for some datasets it may occur that there are training tuples $T_{t1}$ and $T_{t2}$, where the input sequences ${\bf u}_{(-\infty,t]}$ are almost the same as 
(or at least similar enough to produce virtually identical reservoir states $\YH{ t_1 } $ and $\YH {t_2 } $ ) but the target sequence values
${\bf q}_{t1}$ and ${\bf q}_{t2}$ can be significantly different, say for example $1$ and $-1$. The network training result is then the average of both values,
which would be $0$ in the example and far from both target values. Still, the transfer entropy ${\cal I}_{\bf q,u}$ can be very high. So it makes sense to let the network
learn the probabilities $p(q_{t}=1 | \YI{ t } )$ and  $p(q_{t}=-1 |\YI{ t } )$, which we assume are sufficiently close to $p(q_{t}=1 | {\bf u}_{(-\infty,t]})$ and  $p(q_{t}=-1 | {\bf u}_{(-\infty,t]})$.



\begin{figure}[t]
\begin{center}\includegraphics[  width=0.36\paperwidth]{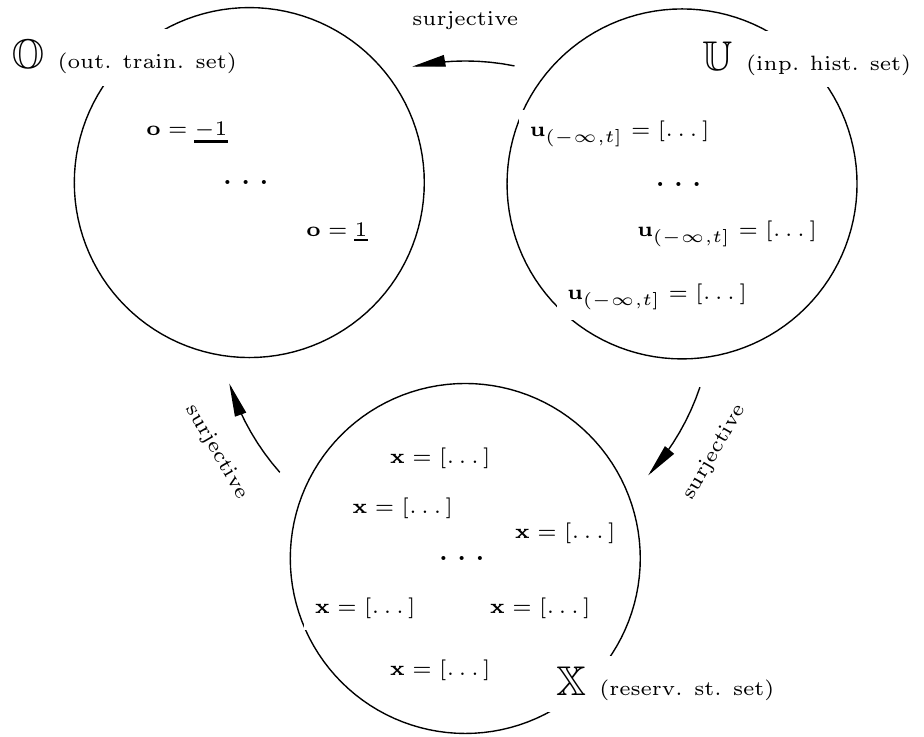}
\end{center}
\caption{\label{setsnrelations} The relations between input sets, reservoir states, and training sets affect the capability of the network to learn. Relations as outlined above
($\mathbb{U} \srel \mathbb{X}$, $\mathbb{X} \srel \mathbb{O}$, $\mathbb{U} \srel \mathbb{O}$) are a necessary prerequisite in order to let the network learn 
$\tilde {\bf o}_t \approx \YO{ t } \in \mathbb{O}$.
}
\end{figure}

Thus, learning processed as in eq. \ref{av_tr} has certain principal limitations. The network can be trained
only if the relation between the input data set and the training output is surjective (cf. fig. \ref{setsnrelations}). 
In other words, it must be possible distribute the set of input sequences into disjunct segments, where every segment refers to a different target.  
In the case of reservoir computing, the relation between the reservoir and the training output has to be surjective. 
The ESP guarantees that the relation between the set of input histories and the set of reservoirs is surjective. Only then the result of the learning can be 
$\YO { t } \approx \tilde {\bf o}_t$. Otherwise, the output is essentially a stochastic estimate with regard to a certain input sequence, that is, one has for every possible output one probability
\[
   p_i = p(\YO{i  } | {\bf u}_{(-\infty,t]}),
\]
where each $ {\bf u}_{(-\infty,t]}$ is an element of the set $\mathbb{U}$ of right infinite input sequences that are different from each other.
As a result after training (cf. the following section) , the trained output becomes
\begin{equation}
   \tilde {\bf o}_t \approx < \YO{  } >_{{\bf u}_{(-\infty,t]}} = \sum_{\YO{ i } } \YO{i } p(\YO{i  } | {\bf u}_{(-\infty,t]}). \label{prob1}
\end{equation}

If the training output consists only of the two possible values $\YO{ 0 } = 0 $ and $\YO{ 1 } =1$, then eq. \ref{prob1} simplifies to 
\[
   \tilde {\bf o}_t \approx p(\YO { } = 1 | {\bf u}_{(-\infty,t]}).
\]

\section{Modeling likelihood estimates and test environment}
\label{sec:likelihoods}

\subsection
{Modeling probability distributions by using the mean square error}
\label{sec:approach}
In the following we outline ideas already used in \cite{2010c,jaeger1} and give some example.
As mentioned above one may consider prediction in the case of a time series.

Instead of training the output with the target function directly, i.e., $\YO{t} = \target{\ti} $, we model a probability that a specific stochastic event may have occurred with regard to the target time series. 
So we define one or a set of events $\omega_r \subset \mathbb{Q}$ and define another type of training scheme,
\begin{equation}
d_{r,t} := \begin{cases} 1 & \mbox{if } \target{\ti} \in \omega_r  \\ 0 & \mbox{if }\target{\ti} \notin \omega_r. \end{cases} \label{train_dt}
\end{equation}

In the scope of this  work the result of eq. \ref{train_dt} is used as a training signal to the network, 
\begin{equation}
o_{r,t} = d_{r,t} 
\end{equation}
The output of a network that has been trained with this kind of training function is an approximation of the probability
\begin{equation}
\tilde{o}_{r,t} \approx p(\target{t} \in \omega_r  |\YI{ [0,t] } ),
\end{equation}
which can be seen from considering the mean square error function
\begin{equation}
E_{MSE} = < \left|(\YO{t} - \tilde{\bf o}_t )^2 \right| >_t = <\left| ({\bf d}_t - \tilde{\bf o}_t)^2 \right| >_t,
\end{equation} 
where $<.>$ is the expectation value operator.
The equation for $E_{MSE}$ can also be written as 
\begin{equation}
E_{MSE} = \sum_r \left[ p(\target{\ti}  \in \omega_r) (1-\tilde{o}_{r,\ti})^2 + (1-p(\target{\ti}  \in \omega_r)) \tilde{o}_{r,\ti} ^2 \right] . \label{MSE}
\end{equation} 

On one hand the 
minimal value of $E_{MSE}$ w.r.t. $\tilde{o}_{r,\ti}$ is reached when
\begin{equation}
\frac{\partial E_{MSE}}{\partial \tilde{o}_{r,\ti}} = \tilde{o}_{r,t}-p(\target{t}  \in \omega_r) = 0,
\end{equation}
so the minimal value of the mean square error as defined in eq. \ref{MSE} is reached if
\begin{equation}
\tilde{o}_{r,t}=p(\hat{\target{t} } \in \omega_r). \label{convproptheo}
\end{equation}

On the other hand, the minimal value of $E_{MSE}$ w.r.t. $\WWHO$ is reached \cite{neurofuzzy1997} when
\begin{equation}
{\bf w}^{out} = (A^T A)^{-1} (A^T B) \label{linreg}
\end{equation} 
where the rectangular matrix $A=[\YH { 0 }, \YH { 1 } \dots \YH { t } ]^T$ and, since ${\bf o}_t = {\bf d}_t$,  $B=[{\bf d}_0, {\bf d}_1 \dots {\bf d}_t]^T$  is composed from the data 
of the training set, and $A^T$ is the transpose of $A$. Note that eq. \ref{linreg} consists of equations that define linear regression.

Thus, after the learning process we can assume
\begin{equation}
\tilde{o}_{r,t} \approx p(\target{\ti} \in \omega_r) \label{convprop}
\end{equation} 
for sufficiently long learning sequences. The result of the learning process is only an approximation (and thus there is a difference between eq. \ref{convproptheo} and eq. \ref{convprop}) for the following reasons.

\begin{itemize}
\item[\sf (L5)\namedlabel{l5}{\sf L5}] The training is over a finite training sequence. Due to the law of large numbers  and due to the finite training set the result of training can only be an extrapolation.
\item[\sf (L6)\namedlabel{l6}{\sf L6}] For situations in the test set that have not appeared in the training set the resulting estimate can only be an approximation.
\end{itemize}
As a result the output of the network can be outside the range between $0$ and $1$. If the output $\tilde{o}_{r,t}$ is larger than $1$, we assume $p(\target{\ti} \in \omega_r)$ to be one. If the output is smaller than $0$ we assume $p(\target{\ti} \in \omega_r)$ to be $0$, repectively.

As a result of common restrictions of reservoir computing (e.g., restrictions above and \ref{l1} -- \ref{l3} as outlined in sect. \ref{sec:esn}), limited information of the input history is encoded in the activity state of the reservoir. Thus, more information about  statistical variables can be retrieved from additional neurons in the recurrent layer.
Since the optimal solution (absolute minimum of the MSE) can be derived, the network is going to find the true probability as far as it is detectable by linear regression from the current state of the reservoir.

In summary, the target of training is that the network after training can map either a one or a zero output to each of the internal states at time $t$ and in such a manner to identify and predict with absolute certainty the output of the network 
to either belong to the event or not. In the case of stochastic time series, it may happen that two different values ${\bf q}_t$ are assigned to identical input time series ${\bf u}_{(-\infty,t]}$ in two different tuples $T_t$. 

In these cases such a deterministic identification is impossible, so the network learns during training to match combinations of zeros and ones to identical internal states
(i.e., virtually the same input histories result in different outputs). The mathematics here is basically the same as the mathematics of solving an over-complete system of linear equations. So the trained output averages out the 
perceived ones and zeros of the trained output, which is identical to a realistic probability estimate.

\begin{figure}[t]
\begin{center}\includegraphics[  width=0.26\paperwidth]{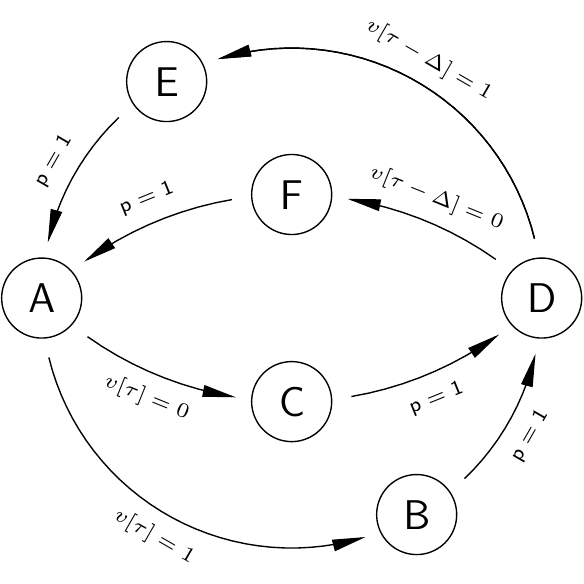}
\end{center}
\caption{\label{testmodel} 
Test model set-up: Depicted is the sequence of states of the test system in order create a time series. 
States $s(t)$ are $\sf A$ to $\sf F$. Transitions between the states 
happen at every time step $t$ and are either completely deterministic
(transitions from $\sf B$, $\sf C$, $\sf E$ and $\sf F$) or ruled by the stochastic array $v[.]$. While the network has no way to infer transitions from state $\sf A$ (since $v[.]$ itself is hidden to the network), 
transitions from $\sf D$ can be inferred from a earlier transition from state $\sf A$. The inference becomes more difficult for the ESN if the value of the constant $\Delta$ is high.
The label $p=1$ indicates deterministic behavior. $\tau$ is incremented each time $\sf A$ is reached. }
\end{figure}

\subsection{Test time series}
\label{sec:experiments}

For the  sake of simplicity, we test here the performance of the model with a test model, where the target time series is the future of the input time series, which is stochastic but has 
an inference scheme.
So, the training output is an item of the input time series in the future, ${\bf q}_t = \YI{ t+1 } $. 

We demonstrate the approach on a prediction task. Our test model cycles between six states $s(t)\in [{\sf A, B, C, D, E, F}]$. Figure~\ref{testmodel} outlines the transition rules. 
The exact shape of the resulting time series is determined by a large array $v[.]$  that is filled with i.i.d random numbers that are either $0$ or $1$  with both having a probability of $0.5$ and a single scalar 
non-negative integer $\Delta$, which in a certain way determines the complexity of the inference. The state transitions work in the following way: 
\begin{itemize}
\item [$\Lambda 0$:] Setting $\tau=-1$
\item [$\Lambda 1$:] Increment $\tau$ by 1
\item [$\Lambda 2$:] Start from state $s(t)={\sf A}$. The value of $v[\tau]$ is read. If $v[\tau]$ is one, then the next state is $s(t+1)={\sf B}$ otherwise the next state is $s(t+1)={\sf C}$.
\item [$\Lambda 3$:] State $s(t+1)={\sf D}$ always follows $s(t)={\sf B}$ and $s(t)={\sf C}$.
\item [$\Lambda 4$:] From state $s(t)={\sf D}$ state $s(t+1)={\sf E}$ follows if $v[\tau-\Delta]=1$, otherwise the next state is $s(t+1)={\sf F}$.
\item [$\Lambda 5$:] Go back to $\Lambda 1$. 
\end{itemize} 
At every iteration the network has access to the current state value $s(t)$. 
The task for the network is then to calculate the probabilities for each possible state to be the next state $s(t+1)$. 
Here the difficulty of inference from different states varies.
The transitions from $B$,$C$, $E$ and $F$ are trivial because
\begin{eqnarray}
\mbox{if } \, \left(s(t)= \, {\sf B}\right) \, \mbox{or} \, \left(s(t)={\sf C} \right) \; &\Rightarrow& \, s(t+1)={\sf D} \nonumber \\
\mbox{if } \, \left(s(t)= \, {\sf E}\right) \, \mbox{or} \, \left(s(t)={\sf F} \right) \; &\Rightarrow& \, s(t+1)={\sf A}
\end{eqnarray}

For the transition from $s(t)={\bf A}$ the network has no way to infer to the next state. The reason is that the network has no access to $v[t]$. So 
\begin{equation}
 s(t)= \, {\sf A} \Rightarrow   \begin{cases} p=0.5 & s(t+1)={\sf B}  \\ p=0.5 & s(t+1)={\sf C}  \end{cases} \label{t0t}
\end{equation}

Finally, the transition from state $\sf D$ can be inferred from an earlier state transition:

\begin{equation}
s(t)= \, {\sf D} \Rightarrow   \begin{cases}   s(t+1)={\sf E} &\mbox{if} \; \; s(t-4\Delta-1)={\sf B}  \\ 
                                               s(t+1)={\sf F} &\mbox{if} \; \; s(t-4\Delta-1)={\sf C}.   \end{cases} \label{t2t}
\end{equation}
Since the memory capacity of the reservoir is limited, the task to reveal the inference of eq. \ref{t2t} is very hard for the network if $\Delta$ has a high value. Thus, for very high values the network falls back to 
the view of the naive observer (one without memory who only is able to infer the next state from the previous one), who assumes

\begin{equation}
\mbox{if} \, s(t)= \, {\sf D} \Rightarrow   \begin{cases} p=0.5 & s(t+1)={\sf E}  \\ p=0.5 & s(t+1)={\sf F}.  \end{cases} \label{t2n}
\end{equation} 

As a general remark, every cycle (from one A to the next A) adds one bit of information to the time series, which should be observed by a system that has a memory of at least $\Delta$ cycles (which would mean that the observer 
would reproduce $v[]$). The information gain in each cycle happens exactly when the network observes the transition from $\sf A$.
Note here that the network's reservoir in no way is adapted to a particular value of $\Delta$. The learning is only done by doing the regression in order to learn the output layer.

With regard to hidden Markov models, one can make two remarks.
A Markov model where every state of the input is represented internally by exactly one hidden state would behave like a naive observer (as in eq. \ref{t2n}) and thus assume the information in each cycle of the time series to be 2 bits. 

In order to design a hidden Markov model with the capability to reveal the relation of eq. \ref{t2t}, at least  $2^\Delta$ hidden states are necessary.
(Compare considerations in \cite{2010c}). 

\subsection{Encoding of states}

The states $s(t)$ of the test model are presented as 6 dimensional vectors in the following way to the network:
\begin{eqnarray}
{\sf A} &\rightarrow& \YI{t} =[ 1, 0, 0, 0, 0, 0] \nonumber \\
{\sf B} &\rightarrow& \YI{t} =[ 0, 1, 0, 0, 0, 0] \nonumber \\
{\sf C} &\rightarrow& \YI{t} =[ 0, 0, 1, 0, 0, 0] \label{input_encoding_eq}        \\
{\sf D} &\rightarrow& \YI{t} =[ 0, 0, 0, 1, 0, 0] \nonumber \\
{\sf E} &\rightarrow& \YI{t} =[ 0, 0, 0, 0, 1, 0] \nonumber \\
{\sf F} &\rightarrow& \YI{t} =[ 0, 0, 0, 0, 0, 1] \nonumber
\end{eqnarray}

In the sense of eq. \ref{train_dt}, one can now consider the 6 following stochastic events for which the network shall calculate the probabilities for the following states. This leads by using the definition of $d_{r,t}$ in eq. \ref{train_dt}. Going through all practical possibilities, one easily can calculate
\begin{equation}
{\bf d}_t = {\bf o}_t = {\bf u}_{t+1}.
\end{equation}

After training and according to eq. \ref{convprop}, the trained outputs are approximately
\begin{eqnarray}
\tilde{o}_{0,t} &\approx& p(\hat{s}(t+1)={\sf A}) \nonumber \\
\tilde{o}_{1,t} &\approx& p(\hat{s}(t+1)={\sf B}) \nonumber \\
\tilde{o}_{2,t} &\approx& p(\hat{s}(t+1)={\sf C})  \\
\tilde{o}_{3,t} &\approx& p(\hat{s}(t+1)={\sf D}) \nonumber \\
\tilde{o}_{4,t} &\approx& p(\hat{s}(t+1)={\sf E}) \nonumber \\
\tilde{o}_{5,t} &\approx& p(\hat{s}(t+1)={\sf F}) \nonumber, 
\end{eqnarray}
where $\hat{s}$ is the best guess of the network.
In the present experiment training was only done when the state $\sf D$ had been reached. The only possible following states were either $\sf E$ or $\sf F$. So 
\begin{equation}
\tilde{o}_{0\dots3,t} \approx 0.
\end{equation}
Moreover, the output units $\tilde{o}_{4,\ti}$ and $\tilde{o}_{5,\ti}$ represent the estimated probabilities that the next state is going to be $\sf E$ or $\sf F$, respectively.
In the following calculations of $\tilde{o}_{4,\ti}$ are discussed, which can be interpreted as a calculation of the network of the probability of $p(\hat{s}(t+1)={\sf E})$. 
The probability for $\sf F$ can be inferred as
\begin{equation}
p(\hat{s}(t+1) = {\sf F}) \approx 1 - \tilde{o}_{4,\ti}.
\end{equation}

\section{MSE simulations}
\label{sec:MSE}
\subsection{Simulation details}
\paragraph{Calculating the output.}
Network training consisting of 3000 iterations is performed after a transient period of 300 iterations.
The output is trained by linear regression where over-fitting was prevented using Tikhonov regularization (ridge regression) with a regularization factor of $\lambda=0.08$. 
In the literature (e.g. \cite{Lokse2017}) one can find examples where the regularization had a significant impact on the performance of the network. Here the purpose was mainly to avoid instabilities in the regression learning. 
Preliminary tests show that learning without regularization nearly have the same results as regularization using $\lambda=0.08$.
After training, the network was tested in period of 3000 iterations. The performance was checked a recording the network predictions for all transitions from state $\sf D$.

\paragraph{Adding noise to the reservoir.}
In order to add noise to the reservoir, the linear response of the network initially outlined in eq. \ref{linear_response} can be modified to
\begin{equation}
\YHLIN{\ti+1} = \WWHH \YH{\ti} +
\beta \WWIH \YI{\ti} + \gamma {\bf \nu}(0,1), \label{linear_response2}\\
\end{equation}
where ${\bf \nu}(0,1) $ is a vector build of normal distributed random values with zero mean and a variance of one and the scalar parameter $\gamma$ indicates the strength of the noise level.

\paragraph{Sampling fitting and extrapolation.}
For each value of $\Delta$ between 0 and 18 and also for different numbers of recurrent neurons, 10 samples were simulated and the average and the standard deviation was taken over all 10 samples for each value of $\Delta$.

\begin{figure}[t]
\centering
\includegraphics[width=0.35\paperwidth ]{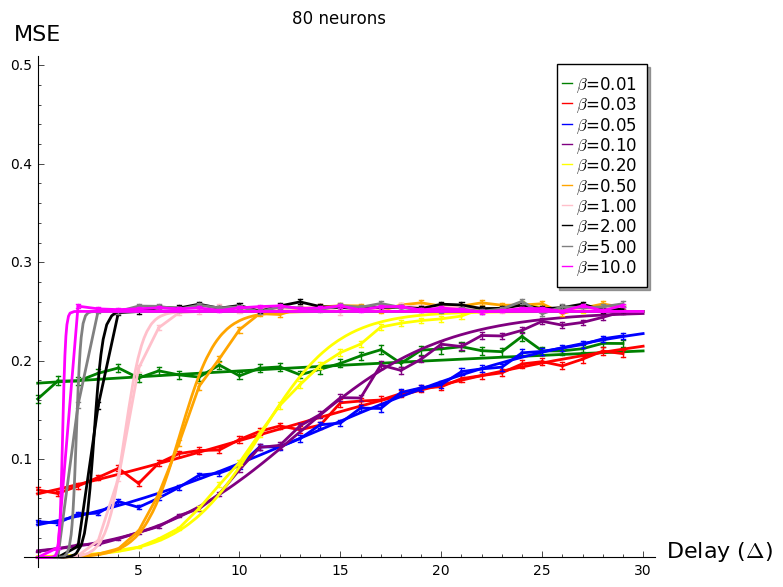}
\includegraphics[width=0.35\paperwidth ]{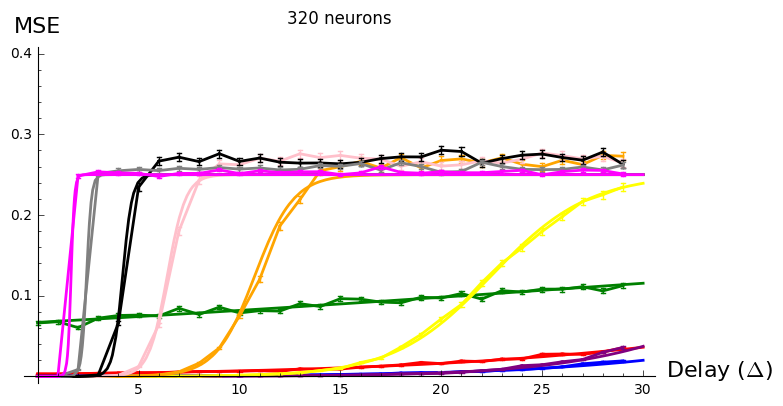}
\caption{\label{results1} 
Top: Errors for different delays. Depicted is the MSE of the probability of the transition to state $\sf E$ in the event of initial state $\sf D$. If the network cannot detect the causality of transitions from $\sf D$, the error is on average 0.25. Different colors indicate different values of $\beta$. The x-axis indicates different delays $\Delta$ of the test system (in cycles of $\tau$, i.e., 4 time steps in $t$). The graph depicts both the real sampled data with error bars and the data fit according to eq. \ref{fitt} for each $\beta$. Network size is in all cases 80 neurons. Bottom: Same plot for a network size of 320 neurons.}
\end{figure}

\subsection{Results}
As outlined above the test sequence is deterministic in the moment of the transition from $\sf D$. Either the next state is $\sf E$ or the next state is $\sf F$. Which one is going to be the next is determined according eq. \ref{t2t}.
So we used the MSE at the transition from state $\sf D$ in order to determine the network performance. A network that is able to infer between the history and the transition to state $\sf E$ can reach zero MSE, whereas for networks that cannot detect the causality the transition appears to be stochastic with equal probability to state $\sf E$ and $\sf F$ (cf. eq. \ref{t2n}). Errors for different delays and different values of $\beta$ are depicted in fig.~\ref{results1}, where both the raw data and the corresponding fits are shown.
 
Here, one can see that for a large range of numbers of neurons in the hidden layer 
a larger number of neurons also results in a better performance of the network.

With regard to $\beta$, one can see that the value of $\beta$ has a significant impact on the MSE for different values of the delay parameter $\Delta$ of the test system. Larger values of $\beta$ result in a better network performance for short delays, whereas lower values of $\beta$ result in lower errors at longer delays\footnote{One possible explanation is that the parameter $\beta$ affects the total activity of the network $|\YH{ t } |_2$. In the case of low activity the non-linear component of the sigmoid transfer function is tiny. Thus, the resulting network is nearly a linear ESN, for which the highest values of memory capacity (MC) -- near the theoretical limit-- 
have already been found earlier~\cite{jaeger1}. On the other hand, the non-linear components are necessary to distinguish sufficiently different vectors in the Hilbert space in order to have a good approximation to model the probabilities.}. For larger network sizes under all conditions, the performance of the network improves accordingly and for all values of $\beta$.

The shape of the data values depicted in fig.~\ref{results1} resemble a sigmoid function, where the left side converges to zero and the right side converges to $0.25$. 

Higher values of $\Delta$ were estimated by extrapolation using the function 
\begin{equation}
\hat{E}_{MSE} (\Delta) = \frac{1}{4+\exp(a \Delta + b)}, \label{fitt}
\end{equation}
where again the parameters $a$ and $b$ were parameters that were fitted in the following way. The function was chosen because it is a smooth sigmoid which appears to fit the data.
The raw data set for each number of neurons and each gain $\beta$ are two arrays that consist of the following fields: 
{\em different delays}($\Delta$), the first array contains the mean MSE (in the following $E_{MSE} (\Delta_i)$, and the second array contains the variance of the mean of the 10 samples that were taken.  From
eq. \ref{fitt} one can calculate
\begin{equation}
\log(1/E_{MSE} (\Delta_i)-4) = a \Delta_i + b, \label{fitt2}
\end{equation}
which gives an equation for every $\Delta_i$. Together they form a system of linear equations, which can be solved by linear regression.

\section{Mutual information between the network and time series simulations} 
\label{sec:info}
\subsection{Simulation details}
In the following, we define the joint probability of the network output and the test system that produces the time series. In particular, we investigate the state transition
from state $\sf D$ to a resulting state of either $\sf E$ or $\sf F$. The result is deterministically determined from an earlier transition from state $\sf A$ to either
$\sf B$ or $\sf C$. In terms of information theory, the mutual information between the previous time series ${\cal S}_t=s_{[0:t]}$ and the next state $s$ at this point (transition from state $\sf D$) 
is one bit. The task of the network is to reveal that relation. One way to measure this ability is to interpret the output of the network at this point in time as
\begin{equation}
p(\hat{s}(t+1)|{\cal S}_t)=\tilde{o}_{4,t}.
\end{equation}
Since --in the test system-- $s(t+1)$ is deterministically derived from ${\cal S}_t$, one formally also can write
\begin{equation}
p(\hat{s}(t+1)|s(t+1))=\tilde{o}_{4,t}.
\end{equation}
That is
\begin{eqnarray}
p(\hat{s}=1,s=1) &=& < p(\hat{s}(t+1)|s(t+1))\cdot \sigma_t >_t, \nonumber \\
p(\hat{s}=0,s=1) &=& < (1-p(\hat{s}(t+1)|s(t+1)))\cdot \sigma_t >_t,  \\
p(\hat{s}=1,s=0) &=& < p(\hat{s}(t+1)|s(t+1))\cdot (1-\sigma_t) >_t, \nonumber \\
p(\hat{s}=0,s=0) &=& < (1-p(\hat{s}(t+1)|s(t+1)))\cdot (1-\sigma_t)>_t, \nonumber 
\end{eqnarray}
where the expectation operator ($<.>$) indicates the average over all occurrences of the state transition from $\sf D$ in the whole time series and $\sigma_t$ is one if $s(t+1)={\sf E}$ and is zero otherwise. 
Mutual information between two random variables is a measure of statistic dependence. One can consider the mutual information 
between the next state of the test system and the time series previous to that state
\begin{equation}
I(\hat{s};{\cal S}) = I(\hat{s};s) = \sum p(\hat{s},s) \log_2 \frac{p(\hat{s},s)}{p(\hat{s}) p(s)},
\end{equation}
in which the mutual information between the current state $s$ of the test system and $\cal S$ is the complete time series until $s$; both $p(\hat{s})$ and $p(s)$ can be assumed to be 0.5. The maximum average value of the mutual information is here 1, the lowest possible value of the mutual information (that of a naive estimator) is zero. The network performance is somewhere in between these two values and depends on the delay parameter $\Delta$ of the test system. Large values of $\Delta$ let the network turn into a naive estimator, while small values of $\Delta$ are expected to result in a better performance of the network.  

For each delay in the range from 0 to 18 and also for different numbers of neurons, 10 samples were taken. $\hat{\cal I}_\Delta$, the fitted value of the mutual information, is defined by 
\begin{equation}
\hat{\cal I}_\Delta = \frac{1}{1+\exp(a \Delta + b)}, \label{fitti} 
\end{equation}
where $a$ and $b$ are fitted parameters again.
They were calculated in a similar way as in eq. \ref{fitt2}. However, the fitting equations were weighted by a weighting function $f({\cal I})$,
\begin{equation}
\log(1/{\cal I} (\Delta_i) -1) \times f({\cal I} (\Delta_i)) = (a \Delta_i + b) \times f({\cal I} (\Delta_i)).
\end{equation} 
The multiplication of $f({\cal I} (\Delta_i)$ does not change the linear equation itself. However, it gives a weight to some equations in case of an over-complete set of linear equations.
Here
\begin{equation}
f({\cal I} (\Delta_i) = \exp(-({\cal I} (\Delta_i)-0.5)^2/0.18),
\end{equation}
so equations that relate to ${\cal I} (\Delta_i) \approx 0.5$ have the largest weight. The weighting was done in order to improve the quality of the fit since the emphasized values are least prone to systematic errors 
and most relevant for the type of S-shape of the curve.

\begin{figure}[t]
\centering
\includegraphics[width=0.38\paperwidth ]{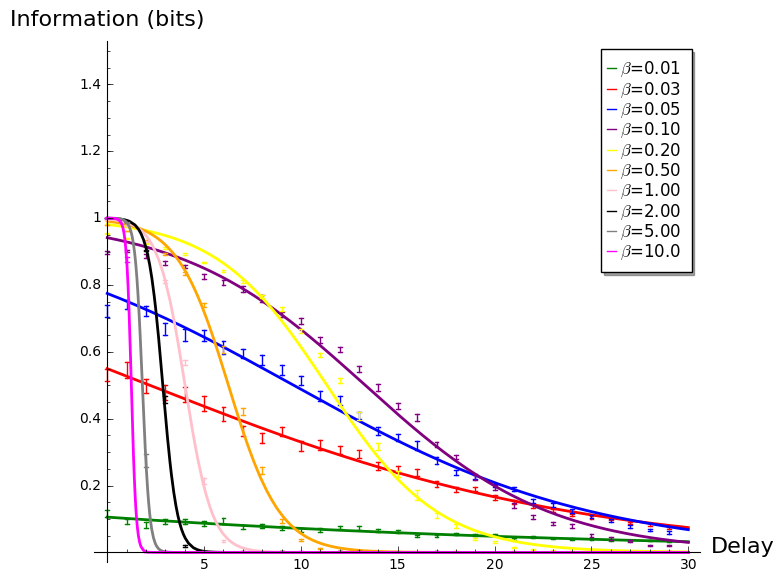}
\caption{\label{results2} 
Depicted are the fits of mutual information between the network output of a network of 160 neurons and the next value of a test system. 
The x-axis shows delay $\Delta$ in cycles $\tau$. Colors encode different values of $\beta$ as detailed in the legend. The corresponding data fit is the curve in the same color.
Average mutual information between the network output and the next state of the test model is always at the state transition from $\sf D$. For short delays the mutual information in almost all cases is near one.
}
\end{figure}

\subsection{Results}  
Fig. \ref{results2} depicts the fits for the network performance of a network of 160 neurons for different values of $\beta$, both the original data (mean and standard deviation) and the curve that was fitted through the data. One can see that in a similar way to the results of Fig. \ref{results1} the performance for short delays is better for higher values of $\beta$ whereas lower values of $\beta$ result in a 
better performance for longer delays.

We define 
\begin{equation}
\hat{\cal I} = \sum_{\Delta \leq \Delta_{max}} \hat{\cal I}_\Delta,
\end{equation}
as an estimate of the total information capacity of each network with regard to the test system, 
where $\Delta_{max}$ was chosen to be 3000. This value $\Delta_{max}$ appeared to be sufficiently large since all considered networks turned into naive estimators in the sense of sect. \ref{sec:experiments} when
$\Delta_{max}=3000$ was used.

\begin{figure}[t]
\centering
\begin{tabular}[t]{ll}
a. & b. \\
\includegraphics[height=0.2\paperwidth ]{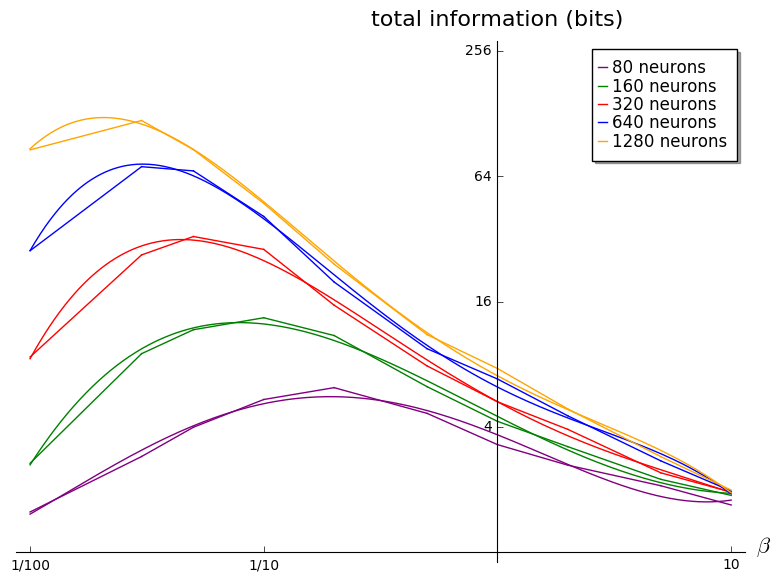} &
\includegraphics[height=0.2\paperwidth ]{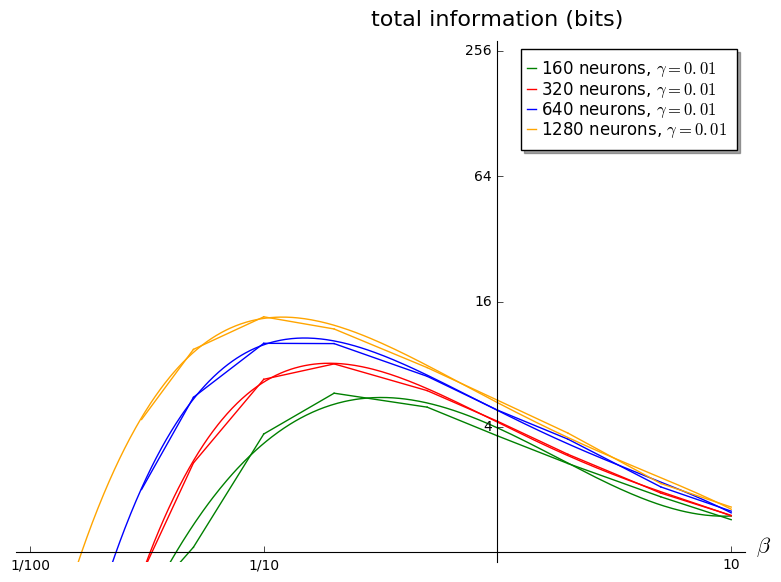}
\end{tabular}
\caption{\label{results3} Overall information capacity of several network sizes. 
{\bf a.} The initial model with different network sizes and $\beta$ factors. The experimental data is plotted together with a 4th order polynomial fit. The maximum level that can be reached with 1280 neurons is about 128 bits. {\bf b.} Adding noise to the activity of the recurrent layer significantly reduces the total information capacity. The plot depicts results of numerical simulations with the same parameters as on the left side but with a noise level $\gamma$ being $10^{-2}$. 
}
\end{figure}

Fig. \ref{results3} shows the resulting estimates of total information content of each of the networks, which vary in their number of neurons and the balance parameter $\beta$. 
The polynomial fits are plotted over the measured data. The pre-factors were again calculated by linear regression.

\begin{figure}[t]
\centering
\includegraphics[width=0.38\paperwidth ]{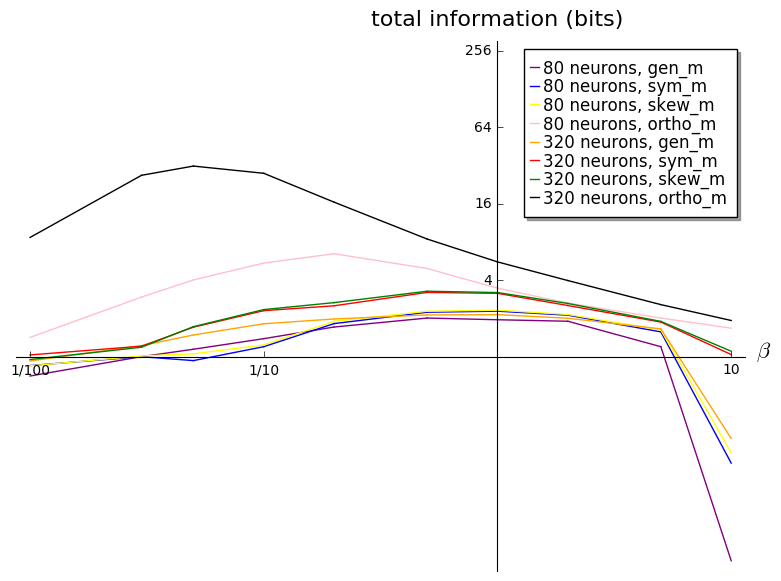}
\caption{\label{results4} 
Same type of simulations are depicted as in fig. \ref{results3} a, except that orthogonal connectivity in the recurrent layer is compared with other types of connectivity. 
General random matrices (labeled as gen\_m), symmetric random matrices (sym\_m), skew-symmetric matrices (skew\_m) are compared. All are normalized to the largest singular values are set to one.
For all types of matrices the sizes 80 and 320 neurons were tested. 
}
\end{figure}

One can see that the optimal balance parameter differs in dependence of the network size. Larger networks show a better overall performance for smaller $\beta$ in comparison to smaller networks. In addition, the network performance 
in the case of noise was tested.  As one can see, under the presence of additional noise the network performance is reduced significantly, in the case of $\gamma=0.01$ to roughly one quarter of the initial performance. Finally, the results of the orthogonal matrices were compared to the performance of general matrices (see fig. \ref{results4}). One can see that for any noise level the performance is much worse than 
for orthogonal matrices. Here the shift of the best memory capacity in dependence of the network size does not exist. Moreover,  the scalability of the networks with orthogonal connectivity is much better than the scalability of the other tested connectivity types. Networks with a general random connectivity matrix do not show any scaling of the performance as a function of size. In the case of the tested networks with symmetric and skew-symmetric matrices, 
we saw a much weaker scalability than for orthogonal networks. So the entropy in those networks only grew by a factor of 1.5 while the network increased by a factor 4. Note that all tested matrices were normalized in a way that the largest singular value was 1.

\section{Discussion}
\label{sec:discussion}

This paper brings together several topics that have not been combined in this way so far but have already been subject to earlier efforts: 
First, the likelihood estimates of time series using reservoir computing done in this approach are similar to \cite{2010c,jaeger1}. Instead of directly estimating the future development of the time series, a probability estimate is calculated. This can be very useful to quantify stochastic time series. Better effects can be achieved with larger networks.
Second, orthogonal recurrent connectivity is used, which has been proposed in several approaches with regard to reservoir computing~\cite{PhysRevLett.92.148102,2010c}. 
Experiments here, like those previous experiments, demonstrate that networks having an orthogonal connectivity pattern show the best performance of all investigated connectivities. Other types of investigated connectivity patterns are much less applicable and have only a fraction of the performance of such networks. Further performance of  networks with orthogonal connectivity scale much better with network size. 
Third, in the case of orthogonal matrices the performance of the network can be significantly modified by changing the absolute strength of the input by varying the parameter $\beta$. 
In other words, one can find a kind of optimal balancing between input connectivity and recurrent connectivity, which has been proposed earlier (e.g. in \cite{verstraeten2010memory}). This is only true for
orthogonal recurrent connectivity. Small values of $\beta$ result in a better reproduction of longer delays $\Delta$, and larger values of $\beta$ result in a better reproduction of shorter values of $\Delta$. Measuring the overall performance of the network over all possible delays shows that the total information transfer shifts for large networks towards smaller values of $\beta$. Fourth, a lower limit of the information content of the network is estimated in bits using Shannon information in a manner similar to \cite{theobi2012}. Initially, J\"ager\cite{jaeger1} proposed a different definition of memory capacity, which measures the effective rank of the network 
by letting it reiterate white noise data. For the future, Shannon information appears to be a more useful measure since here the network behavior with regard to different types of statistics can be 
evaluated more precisely. Finally, we tested the noise sensitivity of echo state networks and how the information processing is impaired by external noise and how orthonormal networks relate in their performance to general random matrices of recurrent connectivity. Here, one can see that noise impairs the the performance of all types of networks significantly. General matrices have a much lower capability than orthonormal matrices.  

Overall, we claim that the present work could be the basis for a new approach to estimate likelihoods of future events based on time series. In this sense the reach of reservoir computing can be extended to statistical estimates of such time series. 

\section*{Compliance with Ethical Standards}
The authors declare that they  have no conflict of interest.

\section*{Ethical approval} 
This article does not contain any studies with human participants or animals performed by the author.

\section*{Acknowledgements}
N.M.M. thanks Oliver Obst for his previous work and the Doctoral Program in Cognitive Sciences at the National Chung Cheng University for setting up an interesting forum for discussions. Earlier preparations for this paper had been funded by the National Science Council of Taiwan and the Ministry of Science and Technology of Taiwan.

\bibliographystyle{plain}
\bibliography{references,references2}

\end{document}